%% file: main.tex
\documentclass[conference]{IEEEtran}
\IEEEoverridecommandlockouts

\AtBeginDocument{%
  \providecommand\BibTeX{{%
    Bib\TeX}}}

\newcommand{\projectname}{{\tt Texture3dgs}\xspace}

\long\def\hidetext#1{}

\usepackage{cite}
\usepackage{amsmath,amssymb,amsfonts}
\usepackage{algorithm}
\usepackage{algpseudocode}
\algrenewcommand\textproc{}
\usepackage{graphicx}
\usepackage{makecell}
\usepackage{subcaption}
\usepackage{textcomp}
\usepackage{xcolor}
\usepackage{xspace}
\usepackage{array}
\usepackage{multirow}
\usepackage{url}
\usepackage{enumitem}
\usepackage{pbalance}

\newcommand{\compactparagraph}[1]{\noindent{\textbf{\textit{#1}}}}
\def\BibTeX{{\rm B\kern-.05em{\sc i\kern-.025em b}\kern-.08em
    T\kern-.1667em\lower.7ex\hbox{E}\kern-.125emX}}

\begin{document}

\title{Optimizing 3D Gaussian Splattering for Mobile GPUs}

\author{\IEEEauthorblockN{Md Musfiqur Rahman Sanim}
\IEEEauthorblockA{
\textit{University of Georgia}\\
Athens, GA, USA \\
musfiqur.sanim@uga.edu}
\and

\IEEEauthorblockN{Zhihao Shu}
\IEEEauthorblockA{
\textit{University of Georgia}\\
Athens, GA, USA \\
Zhihao.Shu@uga.edu}
\and

\IEEEauthorblockN{Bahram Afsharmanesh}
\IEEEauthorblockA{
\textit{University of Georgia}\\
Athens, GA, USA \\
bahram.afsharmanesh@uga.edu}
\and

\IEEEauthorblockN{AmirAli Mirian}
\IEEEauthorblockA{
\textit{University of Georgia}\\
Athens, GA, USA \\
AmirAli.mirian@uga.edu}
\and

\IEEEauthorblockN{Jiexiong Guan}
\IEEEauthorblockA{
\textit{William \& Mary}\\
Williamsburg, VA, USA \\
jguan@wm.edu}
\and

\IEEEauthorblockN{Wei Niu}
\IEEEauthorblockA{
\textit{University of Georgia}\\
Athens, GA, USA \\
wniu@uga.edu}
\and

\IEEEauthorblockN{Bin Ren}
\IEEEauthorblockA{
\textit{William \& Mary}\\
Williamsburg, VA, USA \\
bren@wm.edu}
\and

\IEEEauthorblockN{Gagan Agrawal}
\IEEEauthorblockA{
\textit{University of Georgia}\\
Athens, GA, USA \\
gagrawal@uga.edu}

}



\maketitle 

\baselineskip=1.01\normalbaselineskip 

\input{tex/abstract}

\input{tex/introduction}

\input{tex/background}
\input{tex/sorting}

\input{tex/design}

\input{tex/evaluation}

\input{tex/related}

\input{tex/conclusion}

\input{tex/ack}

\bibliographystyle{ieeetr}
\bibliography{refs,reference}

\end{document}

%% file: tex/abstract.tex
\begin{abstract}
Image-based 3D scene reconstruction, which transforms multi-view images into a structured 3D representation of the surrounding environment, is a common task across many modern applications.  
3D Gaussian Splatting (3DGS) is a new paradigm to address this problem and offers 
considerable efficiency as compared to the previous methods. Motivated by this, and considering 
various benefits of mobile device deployment (data privacy, operating without internet 
connectivity, and potentially faster responses), this paper develops \projectname, an optimized 
mapping of 3DGS for a mobile GPU.  
A critical challenge in this area turns out to be 
optimizing for the two-dimensional (2D) texture cache, which needs to be exploited for faster executions on mobile GPUs. As a sorting method dominates the computations in 3DGS on mobile platforms,
the core of \projectname is a novel sorting algorithm where the processing, 
data movement,  and placement are highly optimized for 2D memory. The properties of this algorithm 
are analyzed in view of a cost model for the texture cache. 
In addition, we accelerate 
other steps of the 3DGS algorithm through improved variable layout 
design and other optimizations. 
End-to-end evaluation shows that \projectname delivers up to  4.1$\times$ and 1.7$\times$ speedup for the sorting and  overall 3D scene reconstruction, respectively -- 
while also reducing memory usage by up to 1.6$\times$ --  
demonstrating the effectiveness of our design for efficient mobile 3D scene reconstruction.


\end{abstract}

%% file: tex/introduction.tex
\section{Introduction}

Image-based 3D scene reconstruction transforms multi-view images into a structured 3D representation of the surrounding environment. This has emerged as a cornerstone technology with  a wide range 
of applications, ranging from  robotics
~\cite{wang2022neural,zhang2024darkgs}, augmented reality (AR)~\cite{kalkofen2008comprehensible,yang2022vox}, to  autonomous systems~\cite{zhou2024drivinggaussian,zhou2024hugs,zhou2024drivinggaussian}.  
This has been a very active area of research --   the success of deep learning models in image-based 3D scene reconstruction  led to the development of novel view synthesis (NVS) models, which can predict novel views of a scene from a set of input images. 
Approaches following this direction,  such as multi-view stereo (MVS)~\cite{furukawa2015multi,kar2017learning,yao2018mvsnet,chen2019point} and implicit methods (e.g., Neural Radiance Fields  or NeRF~\cite{mildenhall2021nerf,barron2022mip,fridovich2022plenoxels,muller2022instant,kerbl20233d}) have shown impressive results 
in generating high-quality 3D reconstructions. However, 
their computational inefficiency leads to high resource requirements. When these methods need 
to support an application on a mobile, edge, or other resource-constrained platform, 
the only practical choice is to use it as a server or cloud for execution~\cite{GSCore}.

3D Gaussian Splatting (3DGS) represents a paradigm shift in scene representation. This approach involves structuring the 
surrounding environment with adaptive and learnable 3D Gaussian primitives, parameterized by spatial position, covariance (defining anisotropic spread), opacity, and view-dependent spherical harmonic color coefficients~\cite{zhou2024feature}.  
Compared to the methods listed earlier   
like Neural Radiance Fields (NeRF)~\cite{mildenhall2020nerf},  
3DGS leverages explicit splatting operations to project Gaussian kernels onto the image plane through differentiable affine transformations. 
This formulation not only has reduced computational requirements, but also inherently supports parallelization, enabling faster rendering speeds while maintaining photorealistic fidelity~\cite{kerbl3Dgaussians}.

The  efficiency of 3DGS naturally leads to its consideration for platforms  such 
as the mobile devices.
On (mobile) device processing of deep learning tasks has been a popular direction in recent 
years~\cite{cai2022enable,niu2020patdnn,niu2022gcd} -- besides advantages such as data privacy and support for operations with low or even no internet connectivity~\cite{huynh2017deepmon,niu2024sod}, mobile device processing can help support latency-critical tasks~\cite{lee2019mobisr,niu2021dnnfusion}. 
The  use of scene reconstruction in applications  such as dynamic obstacle avoidance for autonomous drones or responsive AR applications~\cite{virtualreality}  has latency requirements as low as 20 ms~\cite{yariv2023bakedsdf}. Such a requirement  cannot be met  typically  by sending 
data/query to a server and receiving a response. 

\begin{figure*}[t!]
    \centering
    \includegraphics[width=0.84\linewidth]{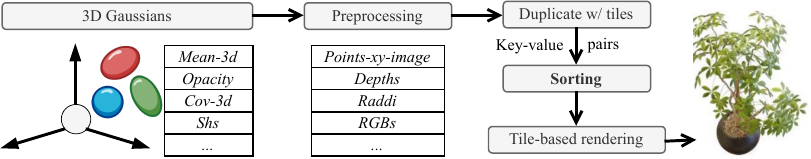}
    \caption{3DGS rendering pipelines.}
    \label{fig:3dgspipeline}
\end{figure*}


The deployment of 3DGS on mobile architectures introduces fundamental challenges due to hardware-software asymmetries. 
Splatting operations such as sorting are inherently memory-intensive, and in comparison, 
mobile memory subsystems are constrained by narrow LPDDR5/X buses (< 50 GB/s bandwidth)~\cite{liang2022romou}.  
Particularly, unlike desktop-level GPUs that typically rely on scratch buffer or shared memory to achieve high performance, mobile GPU applications achieve better performance by exploiting texture memory (and associated cache)~\cite{liang2022romou,guan2025ics}. 
However, the two-dimensional nature of the cache 
requires new approaches to designing algorithms and tuning the implementations. 

There is a line of work on optimizing 3DGS for desktop GPUs, focusing on topics like efficient Gaussian point pruning~\cite{girish2025eagles},  
optimizing memory access patterns~\cite{GSCore}, and efficient rendering~\cite{niemeyer2024radsplat}.
These approaches are either not directly applicable, and/or still leave significant challenges, for the following two reasons: 
1) Mobile GPUs are better suited for tile-based rendering, i.e., have higher memory locality requirements; 
and 2) the limited unified memory hierarchy in mobile GPUs intensifies contention during rasterization of overlapping Gaussians.

Addressing the challenges that we have discussed, this paper makes the following contributions.

\begin{itemize}[leftmargin=*,noitemsep,nolistsep]
    \item We introduce a novel 
    sorting algorithm optimized for GPUs for 2D texture memory. Our method  improves on the limited previous work on sorting  with  texture memory~\cite{Govindaraju2006}  and achieves significantly better 
    cache reuse by careful index transformation, ensuring that 
    pair of values compared (and potentially swapped) in each step are adjacent. 
    
    \item We also design schemes for variable packing and layout organization, in view 
    of the use of different data structures in the entire application, GPU-based parallel 
    processing, and properties  of texture memory.  
    
    \item Further adding a number of optimizations, we implement a complete mobile-optimized 3DGS pipeline, namely \projectname.
\end{itemize}

\projectname has been extensively evaluated on different off-the-shelf mobile platforms, covering representative 3D Gaussian Splatting (3DGS) workloads across various scenes and model complexities. 
Compared to state-of-the-art baseline implementations, our optimized sorting algorithm achieves up to 4.1$\times$ performance improvements by effectively utilizing 2D texture caches. 
Furthermore, the full implementation of \projectname demonstrates up to 1.7$\times$ end-to-end speedup, alongside memory savings of up to 1.6$\times$, 
achieved through efficient variable packing and data layout organization,
highlighting the practical potential of our techniques to enable efficient, real-time 3D scene reconstruction applications on resource-constrained mobile platforms.

%% file: tex/background.tex
\section{Background}


\subsection{3D Gaussian Splatting Rendering Pipeline} 

As stated earlier,  explicit representations like 3D Gaussian Splatting (3DGS) \cite{kerbl3Dgaussians} have emerged as efficient alternatives to prior methods, particularly the Neural Radiance Fields (NeRF) introduced by Mildenhall et al. \cite{mildenhall2020nerf}. 
For scene reconstruction, this method efficiently converts a given camera viewpoint and 3D Gaussian primitives into a rendered 2D image through four main pipeline stages (shown in Figure~\ref{fig:3dgspipeline}): 

\noindent{\bf Preprocessing.}
In this stage, the camera parameters and 3D Gaussian properties, i.e., position (mean), opacity, covariance matrix, and spherical harmonics, are processed to calculate the visual attributes of each Gaussian. A step called {\em Frustum culling}  eliminates Gaussians outside the camera view. The remaining Gaussians are projected onto the 2D image plane, forming elliptical footprints with updated attributes necessary for rasterization.

\noindent{\bf Duplication with Tiles.}
The image plane is subdivided into 16$\times$16 tiles. Each projected 2D Gaussian ellipse is represented by an axis-aligned bounding box (AABB). Gaussians are duplicated across all tiles that intersect their AABB. These duplicates form {\em key-value } pairs, indexed by a combination of the tile identifier and Gaussian depth (i.e., distance from the camera).

\noindent{\bf Sorting.}
The duplicated 2D Gaussians are first sorted based on tile indices to group Gaussians belonging to the same spatial region. Within each tile, Gaussians are further sorted by their depth values to ensure correct rendering order for transparency and occlusion. 

\noindent{\bf Rendering.}
During the final rendering step, the color and opacity values from sorted 2D Gaussians are composited through a step called {\em alpha blending}, determining the final color of each pixel within a tile. This structured pipeline leverages GPU rasterization for efficient, high-quality, real-time rendering of complex scenes.



\subsection{Mobile GPUs and Texture Memory} 

\begin{figure}[t!]
    \centering
    \includegraphics[width=.92\linewidth]{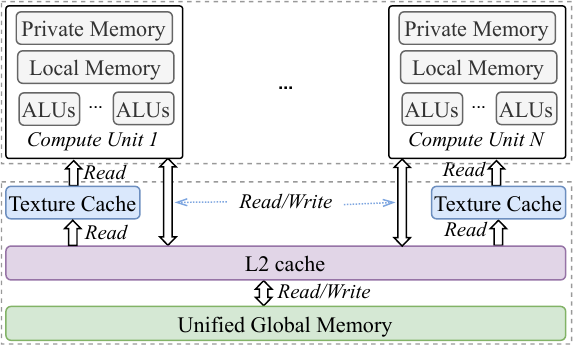}
    \caption{Mobile GPU memory hierarchy.} 
    \label{fig:mobilegpu} 
\end{figure}

Modern mobile GPUs utilize a specialized memory hierarchy optimized for energy efficiency and reduced memory bandwidth usage~\cite{smartmem}. 
Unlike desktop GPUs, mobile GPUs primarily employ a  2.5D texture memory, with a corresponding 
read-only cache  (Figure~\ref{fig:mobilegpu}). The 2D part of the memory signifies support for spatial 
locality along two dimensions, while the .5 part implies that each {\em texture element} (also called 
a {\em texture point})  is actually a 
vector of length 4.  
Two key design aspects critically affect texture memory performance: texture representation and cache organization~\cite{hakura1997design, igehy1998prefetching, doggett2012texture}. Unfortunately, both are often patented and not completely understood  for mobile platforms~\cite{qualcomm2023}. Texture representation refers to how 2D texture data is stored in memory, including the {\em data unit} (e.g., {\em 2D blocks}), their {\em layout} (e.g., hierarchical blocking), and the {\em storage order} (e.g., row-major, column-major, zigzag, or Hilbert). The cache organization covers how data is fetched between the texture cache and main memory, including mapping, replacement policies, and other hardware-specific strategies.
The support for 2.5D texture memory significantly reduces memory bandwidth requirements but imposes constraints on 
how memory accesses should be optimized. 
More specifically,   adapting algorithms such that their data access patterns match the 
properties of the texture cache can be a significant challenge. 

%

\subsection{Sorting on GPUs}  

Sorting is a fundamental and one of the most widely studied problems in computer science. 
Traditional CPU-based sorting algorithms suffer from significant performance limitations on GPUs due to inadequate parallelism and frequent cache misses.  Researchers have explored GPU-based sorting, leveraging its massive parallel processing capabilities by adapting sequential sorting algorithms into suitable parallel implementations \cite{parallaelsort, UberFlow, gputerasort, cudaquick}.
Early GPU-based sorting algorithms were predominantly derived from sorting networks, such as Batcher’s bitonic and odd-even merge sort\cite{bitonic} and Dowd’s periodic balanced sorting networks \cite{pbsn}, the latter being inspired by the bitonic sort network. While the bitonic sort network is well-suited for parallel processing,  
it does have a high memory access complexity of \( O(n \log^2 n) \).  

Most recent research in GPU-based sorting has primarily focused on NVIDIA GPUs and CUDA-based implementations 
(and largely targeting  integer arrays)~\cite{cudabitonic, cudaquick, radix}.
With advancements in GPU architectures and the introduction of shared memory, sorting strategies evolved. Instead of applying a global sorting network, modern approaches first partition the sequence into sub-sequences, which are independently sorted in parallel. These sorted sub-sequences are then merged in parallel to construct the final sorted sequence~\cite{gpuquicksort, effmergesort}. For a comprehensive overview of advancements in parallel sorting techniques, readers can refer to the surveys in \cite{owens2007survey, singh2018survey}. 

%% file: tex/sorting.tex
\section{Texture Cache Friendly Sorting  }

\subsection{Sorting on Texture memory} 
There is very little work on optimizing sorting for mobile texture memory. 
The original approach by Govindaraju et al.~\cite{govindaraju2005cache} targeted texture 
memory/cache on desktop GPUs. This work was later extended into GPUTeraSort~\cite{gputerasort}, 
which we will carefully examine below. 
Since then, there has been minimal development of GPU sorting algorithms specifically optimized for mobile architectures. Notably, popular mobile deep learning frameworks, including MNN~\cite{alibaba2020mnn} and NCNN~\cite{Ni_ncnn_2017}, do not offer specialized sorting operations 
for texture memory.


\begin{figure}[t!]
    \centering
    \includegraphics[width=0.92\linewidth]{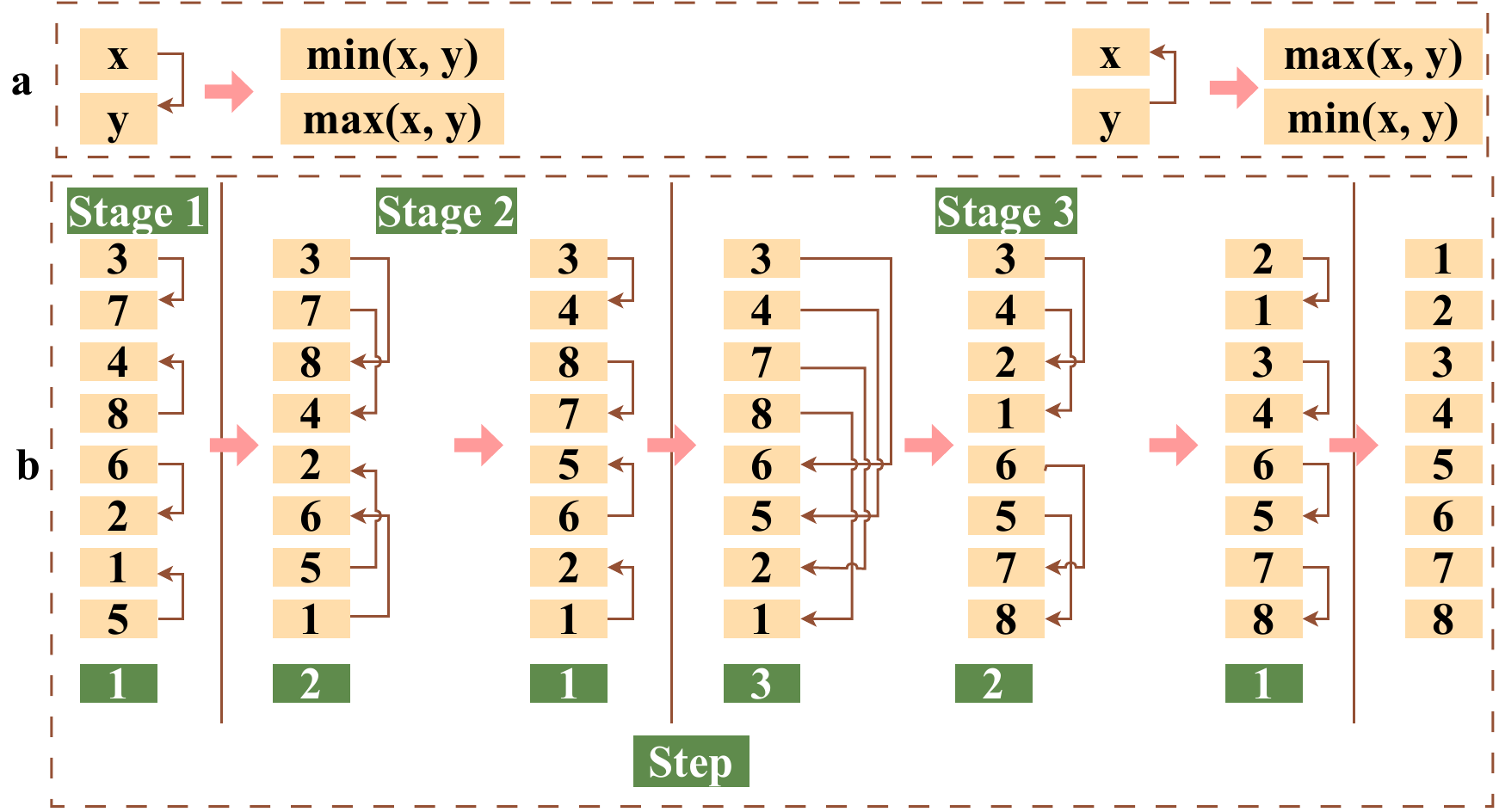}
    \caption{Bitonic sorting network illustration: \textmd{(a) comparator operations, where two elements are compared and swapped based on their order, and (b)   sorting process across multiple steps, where at each stage, the array is conceptually partitioned into sorted segments (length of \(2^{step}\)). The arrows indicate the comparisons and exchanges at each step. }}
    \label{fig:bitonic-sort} 
\end{figure}

GPUTeraSort leveraged Dowd’s Periodic Balanced Sorting Network (PBSN)~\cite{pbsn} to achieve improved memory usage while sorting large key-value datasets. 
As background for presenting the existing work and its improvements toward our algorithm, we review the well-known bitonic sorting 
method. 
The key concept here is a {\em bitonic sequence}, which is a sequence that is either entirely non-decreasing or non-increasing. The algorithm starts with the input array \(a = (a_0, a_1, \ldots, a_{n-1})\) and works from the bottom up, gradually merging smaller bitonic sequences of equal size. Initially, pairs of adjacent elements, like \((a_{2i}, a_{2i+1})\), are merged to form bitonic sequences of size 2. In the next stage, these smaller sequences are merged into larger ones of size 4, and this continues with the size doubling at each stage. To obtain a fully sorted list, we require \( \log(n) \) stages. Figure \ref{fig:bitonic-sort} illustrates the full bitonic sort network process.

\begin{algorithm}
\caption{GPUTerasort's Bitonic sort process}\label{algorithm:gputerasortbitonic}
\begin{algorithmic}[1]
\Procedure{BitonicSort}{$texture$, $W$, $H$}
\State $n \gets$ numValues to be sorted $ \gets W \times H $ \Comment{single array representation}
\For{$i = 1 \to \log n$} \Comment{for each stage}
    \For{$j = i \to 1$}
        \State Quad size $B = 2^{j - 1}$
        \State Compare and swap within the Quads of size $B$
        \State Copy from frame buffer to $texture$
    \EndFor
\EndFor
\EndProcedure
\end{algorithmic}
\end{algorithm}

\begin{figure}
    \centering
    \includegraphics[width=0.5\linewidth]{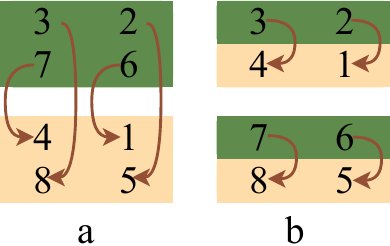}
    \caption{GPUTeraSort's sort \textmd{a) stage 2,  step 2 and b) stage 2,  step 1 -- the quad sizes 
    are 4 and 2, respectively.  During each comparison step, a value from the green region is paired with its corresponding value from the yellow region. After comparison,   the minimum and maximum values are placed in the corresponding green and yellow positions,  respectively. }}
    \label{fig:sort_stage} 
\end{figure}

GPUTeraSort, which maps this process to GPUs with texture memory, is summarized as Algorithm~\ref{algorithm:gputerasortbitonic}.  
If \( n \) is the number of values to be sorted, the algorithm proceeds in 
\(log(n)\) {\em stages}. In the \( i^{\text{th}} \) stage, there are \( i \) {\em steps}, executed in reverse order from \( i \) down to \( 1 \). Each step is responsible for building and merging bitonic sequences of size \( 2^i \), progressively combining smaller sorted sequences into larger ones until the entire array is sorted. 
The algorithm operates on texture memory by reading 
from an input texture and writing to an output texture. The output texture 
of one stage becomes the input texture for the next stage, while the former input texture is now free to store the new 
output. 
The key concept here is the {\em quad size}--specifically, in {\em compare 
and swap} using a quad of size \( B \), a location from one quad is compared (and potentially swapped) with a corresponding location in the next quad (with specific pairing chosen to minimize divergence). This process is illustrated in 
Figure~\ref{fig:sort_stage}. Each of these quads is processed in a separate kernel to 
achieve parallelism.

To understand how these quads are physically allocated in 2D memory with optimal layouts, 
with \( n \) being the number of values that need to be sorted, 
the texture dimensions \( W \) and \( H \) are set to powers of 2 that are 
each closest to \(\sqrt{n}\) (and such that \( W \times H \) is closest to \( n \)).   
When working with a quad size of \( B \), the 
texture is segmented as follows. Each quad occupies a continuous dimension 
when \( B < H \); otherwise, the quad forms a block with dimensions 
\( H \) and \( B/H \). 
The implications of this allocation will be discussed 
after we present a cost model for texture memory.

Several details of  texture cache architecture, 
such as specific replacement policies or exact block mappings, 
are typically proprietary. 
Our algorithm relies  on a precise offline profiling (as detailed in the next section) involving micro-benchmarking and modeling  of cache performance to accurately capture mobile GPU characteristics. 
This approach enables us to approximate hardware behavior effectively 
ensuring robustness across multiple mobile GPUs (as validated in our Section~\ref{sec:eva-portability}).

\begin{table}[]
\centering
\caption{Weights in machine learning model developed for relating two-dimensional strides to latency.}
\label{tab:histogram}
\begin{tabular}{|l|l|l|l|l|l|} \hline 
Block Size  & 2  & 4  & 8  & 16 & 32   \\ \hline 
Cross Horizontal Block  &  0.64 & 0.03   & 0.26   & 0.55 & 0.40   \\ \hline 
Cross Vertical Block  & 0.81   & 0.62   & 0.87   &  0.29  & 0.4   \\ \hline 
\end{tabular} 
\end{table}

\input{tex/cost.tex}

\begin{figure}
    \centering
    \includegraphics[width=1\linewidth]{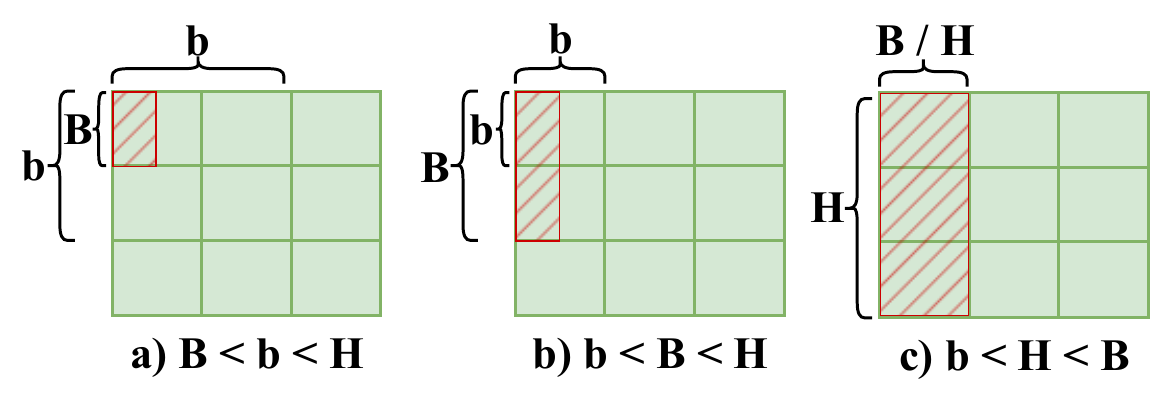}
    \caption{GPUTeraSort’s texture memory allocation \textmd{for different quad sizes: $b$ is the texture cache block size,  $B$ is the quad size, $W$  and $H$ are texture dimensions. This figure shows three cases: (a) quad is along a single dimension, with length less than $b$, (b) the quad is along a single dimension, with length greater than $b$, and (c) the quad is two-dimensional with 
    length $H$, and width $B/H$. }}
    \label{fig:block} 
\end{figure}

 \subsection{Optimizing Texture Memory Based Sorting }
 
It is easy to see that the original TeraSort algorithm is not optimized for the modern texture cache. 
If the quad size is \( B \), the distance between two compared pixels turns out 
to be an average of \( B \), as shown in Figure~\ref{fig:sort_stage}. 
While GPUTeraSort optimizes data access when \( B < b \), since all comparisons happen within 
a single texture cache block (Figure~\ref{fig:block}\textit{a}), 
this advantage diminishes when \( B \geq b \) (Figure~\ref{fig:block}\textit{b} and \textit{c}), 
because the values being compared are in different cache blocks. This leads to a large number of misses in the texture cache. 
In addition, it should be noted that, because of the read-only nature of the texture cache, the initial reading of a quad always causes cache misses.

In view of this analysis, we now present a sorting algorithm optimized for texture memory on modern mobile GPUs. 
The key idea in this work is a {\em layout transformation}, 
which we explain first.

\begin{figure}[t]
    \centering
    \includegraphics[width=0.9\linewidth]{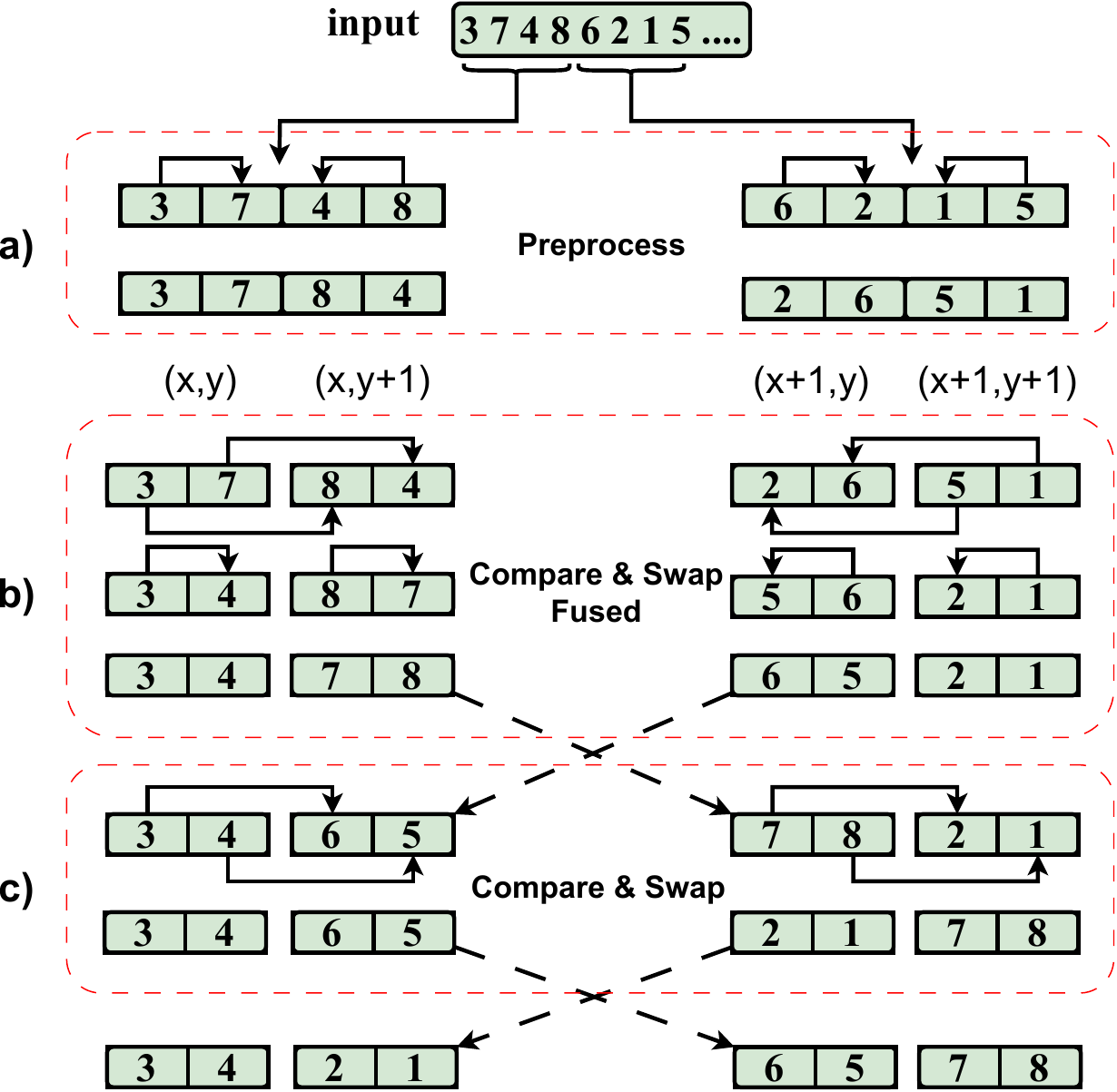}
    \caption{Sorting pipeline: \textmd{(a) Stage 1 comparison and swap operations; (b) Stage 2, steps 2 and 1 comparison and swap operations—all comparisons are with immediate vertical (step 2) or horizontal (step 1) neighbors only (the same process occurs in Stage 3, steps 2 and 1 as well); and (c) Stage 3, step 3 -- all comparisons and swaps are with immediate vertical neighbors only. Solid arrows represent compare-and-swap operations, and dotted arrows represent data movement.}}
    \label{fig:sort} 
\end{figure}

\noindent 
{\bf  Layout Transformation.}
Our sorting algorithm is designed to take full advantage of the texture cache's features by keeping the comparing pairs for each stage physically close in memory. 
This concept is illustrated in Figures~\ref{fig:sort}. 
The core idea is as follows: if a texture point \textbf{\textit{a}} at coordinate \((x, y)\) is to be compared with another point \textbf{\textit{b}}, then \textbf{\textit{b}} is placed at either 
\((x, y+1)\) or \((x+1, y)\). 
To make this possible, we apply a layout transformation at every sorting step. 
This layout transformation is applied while the output tensors are written. 
As a result, the layout transformation does not cause any additional data movement, though there can be a cost associated with index transformation-related computations. 
This layout transformation is feasible because of the predefined structure of the bitonic sorting network.

\begin{figure}[t]
    \centering
    \includegraphics[width=0.92\linewidth]{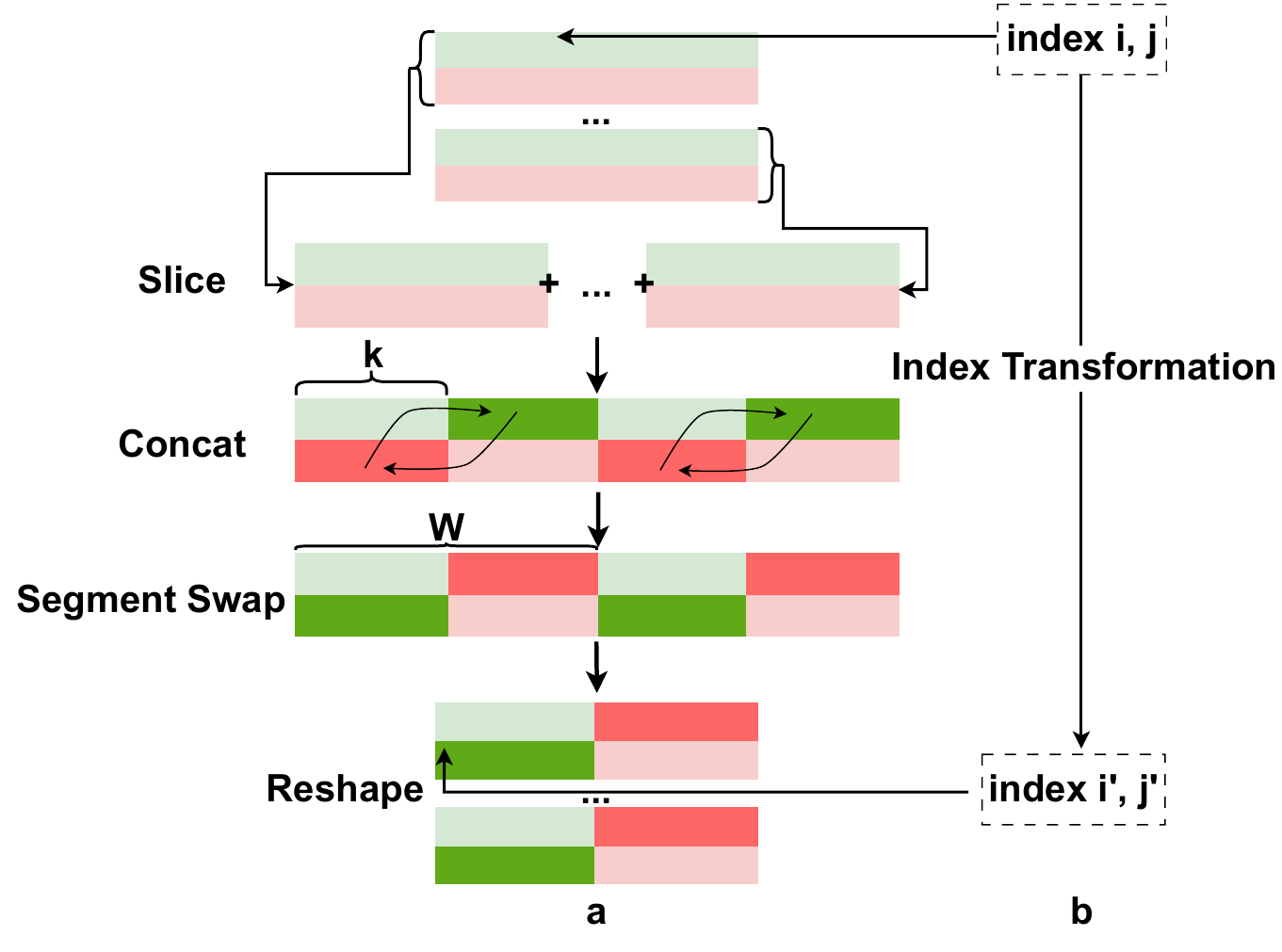}
    \caption{Layout transformation (logical view).  \textmd{$k$ is segment length and $W$ is the original texture width dimension. Even and odd parts are shown in green and red, respectively, with swapped segments highlighted.}
    \label{fig:layout_trasformation}} 
\end{figure}

The concept behind the index transformation can be shown through 
four steps in Figure~\ref{fig:layout_trasformation}, i.e., {\tt Slice}, {\tt Concat}, {\tt Segment Swap}, and {\tt Reshape}.  Initially, the texture is {\em sliced}  by grouping every two consecutive rows and concatenating each set 
of (odd or even) rows. As a result, all even-numbered rows are {\em concatenated} into the first group, and all odd-numbered rows into the second group.
Next, each of these groups is divided into {\em  segments}, where the segment size is determined by the quad size at the next step, specifically,   \( k= 2^{\text{step} - 2} \).  
Following this, a {\em segment swap}  is performed between the two groups --  the odd-numbered segments from the first group  are swapped with the even-numbered segments from the second group, i.e., we perform the operation:

\begin{IEEEeqnarray}{c}
\text{swap}(first\_group\_segment_i, second\_group\_segment_{i-1}) \nonumber \\
i \in \{1, 3, 5, \ldots\}
\end{IEEEeqnarray}

After the segment swap, the elements of the group are reshaped back to their original texture width and height.  
This reshaping ensures that the comparing pairs are positioned correctly for the next sorting step. 
In the resulting layout, every element at coordinate \((x, y)\) in an even-numbered row is directly aligned with its comparison partner at \((x, y+1)\), aligning perfectly with the comparisons needed at this step in the original bitonic sort network.   
It is important to emphasize that the above four steps are conceptual, intended to present the method. 
For actual implementation, we apply an {\em index transformation} along with the compare-and-swap operation of 
the previous step. 
This results in the placement of each value into its correct position for the next sorting step.  
Although this introduces some additional index computation costs, the overhead is minimal compared to the performance gain from improved cache efficiency, which we 
analyze later. 
This trade-off is especially favorable in the context of GPU sorting, where performance is typically bound by memory accesses rather than computations.



\noindent 
{\bf  Stage Fusion.} 
Another optimization targets the fact that each texture point has 4 values. 
For each stage of the bitonic sort, each kernel thread reads two textures, each containing two key-value pairs, for a total of four key-value pairs. In the last two stages, only these four elements are required for that particular step. Therefore, instead of using two separate kernels for the last two stages, we can combine them into a single kernel.  Given that the four elements share the same vector, this reduces the total 
memory traffic.   

\begin{algorithm}[H] 
\begin{small} 
\caption{Sorting Pipeline}
\label{algorithm:pipeline}
\begin{algorithmic}[1]
\Procedure{SortingPipeline}{$keys, values$}
    \State $n \gets \textit{array size of keys}$
    \State $x \gets ceil(log_2(n))$ 
    \State  Choose dimensions $H$ and $W$ 

    \State $tex[2][W][H] \gets init$
    \State $texPoint \gets 0$
    \State $PreprocessKernel(keys, values, tex[texPoint])$
    \For{$i = 2 \to x$} \Comment{stage}
        \For{$j = i \to 3$} \Comment{step}
            \State $CompareSwap(tex[texPoint], tex[1-texPoint], i, j)$
            \State $texPoint \gets 1 - texPoint$
        \EndFor
        \State $CompareSwapFused(tex[texPoint], tex[1-texPoint], i)$
        \State $texPoint \gets 1 - texPoint$
    \EndFor
\EndProcedure
\end{algorithmic} 
\end{small} 
\end{algorithm}

\noindent 
{\bf Overall Algorithm.} 
With these two key novel optimizations, the overall sorting method, 
presented as   Algorithm~\ref{algorithm:pipeline} operates as follows. 
The process begins with the {\textit{PreprocessKernel}} (Line 7), which prepares the input data by organizing it into a newly structured texture layout and performing the first stage of bitonic sorting. This initial setup is detailed in Figure~\ref{fig:sort}(a). Each  {\textit{PreprocessKernel}} thread reads 4 keys and 4 values as vectors of 4 -- this size supports SIMD processing. 
The pipeline then continues through the remaining \( (x - 1) \) sorting stages (Line 8). In the 
\( i^{\text{th}} \) stage, a total of \( i \) steps are performed: the first \( (i - 2) \) steps are executed using the  {\textit{CompareSwap}}  while the final two steps are fused into a single pass using the {\textit{CompareSwapFused}} (Algorithm~\ref{algorithm:pipeline}, Line 13).  
The method  {\textit{CompareSwap}} (Algorithm~\ref{algorithm:kernel}) reads two row-adjacent texture points, determines their sorting order according to the bitonic sorting network, swaps them if necessary, computes the updated indices,  and writes them back to the output texture.  As stated previously, the last 
two stages are merged together -- the primary difference between \textbf{{\textit{CompareSwap}}} and \textbf{{\textit{CompareSwapFused}}} is that the latter performs an additional {\em intra-texture point} comparison step. 

\begin{algorithm}[H]
\caption{Kernel for Sorting with Texture Memory}
\label{algorithm:kernel}
\begin{algorithmic}[1]
\Procedure{ComapareSwap}{$inTex$, $outTex$, $stage$, $step$}
    \State $upper\_point \gets inTex[x][y << 1]$
    \State $lower\_point \gets inTex[x][y << 1 + 1]$
    \State $dir \gets get\_direction(x, y)$
    \State $compare\_and\_swap(upper\_point, lower\_point, dir)$
    
    \State \text{Calculate index for next stage and write to } $outTex$
\EndProcedure
\end{algorithmic}
\end{algorithm}

\noindent 
{\bf Cost Analysis.} 
Our advantages arise because of both layout transformation and stage fusion. 
Due to this fusion, our approach significantly reduces the number of memory accesses compared to the baseline texture-based sorting algorithm, GPUTeraSort \cite{gputerasort}. The total number of memory accesses required by our algorithm for sorting is given by \(n \log(n) \times \frac{\log(n) - 1}{2}\). In contrast, the total memory accesses required by Govindaraju et al. \cite{govindaraju2005cache}:\(5 \times n \log(n) \times \frac{\log(n) + 1}{2}\) because they do not have any stage fusion.
In terms of texture cache hits and misses -- as discussed earlier, the L1 texture cache is optimized for 2D spatial locality, meaning that data stored in a 
square block of dimension size  \( b \) can be efficiently accessed with minimal cache misses. Given a texture of width \( W \) and height \( H \), the theoretical minimum number of cache misses is for any computation that traverses all the data is $
\frac{W \times H}{b^2}$.  In our sorting kernel, each comparison requires two data reads, and as explained, these two data points should be close.  Thus, unlike the original algorithm, we achieve a 
cost close to the minimum. 


%% file: tex/cost.tex
\subsection{Cost Model for Texture Memory Based Processing} 


For predicting the performance of an algorithm that accesses the texture cache, we 
refer to a recent study~\cite{guan2025ics}. As background, in a 2D texture cache, data is typically organized in {\em 2D blocks}, enabling 
data locality along both the width and height dimensions~\cite{doggett2012texture}. 
However, because the exact architectural details are not publicly known, this empirical study leveraged a set of micro-benchmarks involving random accesses to  
establish the relationship between data accesses and memory latency for a single thread.

In this process, 
inspired by pointer-chasing benchmarking~\cite{volkov2008benchmarking}, two-dimensional random access indices were generated offline.
This step takes a list of possible strides as input and constructs a multinomial distribution, 
with each benchmark execution assuming a different multinomial distribution to ensure randomness. 
Next, a micro-benchmark kernel was used 
to measure memory latency values for a set of random strides. 
This micro-benchmark kernel operates in a pointer-chasing style, i.e., it fetches a pixel and uses its value as 2D strides for subsequent accesses.

Next, the concept of {\em cross-block stride} is introduced --
it is defined as a function of the shape and size of the data block, 
and is {\em the stride that  
goes across distinct data blocks} (under the 
assumed shape and size of the block).
The benchmarking process collects the cross-block strides for various assumed data block shapes and sizes, along with the execution latency for each run. 
This information is used as the input feature for the subsequent machine learning model. 
This leads to the training data \((H, L)\), where \(H\) is the histogram of cross-block strides for the assumed data block shapes and sizes, and \(L\) is the profiled latency for each run. 
The collected training data are fed into a machine learning regression model based on the least squares method.
 


The results have shown that the performance can be effectively captured in terms 
of cross-block stride-based summarization of data accesses~\cite{guan2025ics}. In one of the 
experiments, the values obtained are shown in Table~\ref{tab:histogram}.   
These results show that, on 
average, there is not a large difference between horizontal and vertical 
accesses. Moreover, the latency can be fully captured by considering
horizontal or vertical strides of up to 32.  
For the analysis of cache performance of algorithms on texture memory, we can assume an abstract two-dimensional 
block size \( b \), such that ``cache misses'' occur when a stride crosses the two-dimensional block.

%% file: tex/design.tex
\section{Design of \projectname} 

This section outlines the key optimizations in our \projectname implementation. Building on the sorting algorithm discussed earlier, we detail how it has been further optimized for use within \projectname.

\subsection{Variable Packing to Exploit Texture Memory}

\begin{table}[t!]
\caption{Grouped parameter input of preprocessing.}
\label{tab:single_gauss}
\centering
\resizebox{\linewidth}{!}{ 
\begin{tabular}{|l|l|c|c|}
\hline
\textbf{Group} & \textbf{Parameter Name}      & \textbf{Number of data points} & \textbf{Number of Texture} \\ \hline
\multirow{2}{*}{Group 1} 
               & Mean3d                 & 3              & \multirow{2}{*}{1}     \\ \cline{2-3}
               & Opacity                & 1              &                        \\ \hline
Group 2        & Cov3Ds         & 6              & 2                      \\ \hline
Group 3        & Shs                    & 48             & 12                     \\ \hline
\end{tabular}
}

\end{table}

\begin{table}[t!]
\caption{Comparison of parameters across operations.}
\label{tab:parameters_comparison}
\centering
\resizebox{\linewidth}{!}{ 
\begin{tabular}{|l|c|c|c|}
\hline
\textbf{}                     & \textbf{PrefixSum} & \textbf{DuplicateWithTile} & \textbf{Render} \\ \hline
Points\_xy\_image    &                    & Yes                        & Yes                    \\ \hline
Depths              &                    & Yes                        & Yes                    \\ \hline
Raddi              &                    & Yes                        & Yes                    \\ \hline
Conic\_Opacity            &                    &                            & Yes                    \\ \hline
RGBs                         &                    &                            & Yes                    \\ \hline
Tile\_touched                           & Yes                &                            &                        \\ \hline
\end{tabular}
}
\end{table}

As we discussed while presenting our sorting algorithm,  designing an appropriate data layout that adheres to the dimensional limits of texture memory is crucial to ensure  efficiency during texture read operations.   
To this end,  the width and height of a texture should be configured to approximate a square shape. 
The key data structure in the code is the set of Gaussians -- each Gaussian is represented by 58 data points, as detailed in Table \ref{tab:single_gauss}. During the preprocessing stage of 3D Gaussian rasterization, these data points must be read and processed, resulting in 12 data points  being output  for subsequent kernel operations (please see Table \ref{tab:output_preprocess}). 

\noindent 
{\bf Input Data Organization.} 
A number of possibilities can be considered for the layout of Gaussian-related 
parameters in texture memory. 
 One possibility is storing parameters for different Gaussians sequentially across the four texture channels. However, with these alignments, as each thread is assigned to 
 process a single Gaussian,  a significant amount of unused data is read. 
 Another possibility  
 is assigning a single thread to process multiple Gaussians to fully utilize the texture data being read -- 
 however, this would reduce the total amount of parallelism in the implementation. 
As a result, we utilize the grouping strategy outlined in Table \ref{tab:single_gauss}, which 
reduces the total number of read operations from 58 to 15. As Group 1 shown in  Table \ref{tab:single_gauss} contains all the necessary information for the parameters of a single Gaussian, the texture array size should match the number of Gaussians. For Group 2, each Gaussian requires 2 texture points to represent its parameters, so the texture size must be twice the number of Gaussians.

\noindent 
{\bf Output Data Grouping.}
Output data grouping  presents additional challenges due to its usage across multiple kernels.  
Table \ref{tab:parameters_comparison} outlines the specific usage of each parameter across future kernels. Based on this analysis, {\tt Points\_XY\_Image}, {\tt Depth}, and {\tt Radii}  are grouped together as they are predominantly accessed during the \textit{DuplicateWithTile} operation. This grouping minimizes the read operations required for that kernel.  
Similar considerations inform the rest of the design of the packing strategy.

\begin{table}[t!]
\caption{Grouped parameter output of preprocessing step.}
\label{tab:output_preprocess}
\centering
\resizebox{\linewidth}{!}{ 
\begin{tabular}{|l|l|c|c|}
\hline
\textbf{Group} & \textbf{Parameter }        & \textbf{No.  of data points} & \textbf{Texture Points} \\ \hline
\multirow{3}{*}{Group 1} 
               & Points\_xy\_image         & 2              & \multirow{3}{*}{1}     \\ \cline{2-3}
               & Depths                   & 1              &                        \\ \cline{2-3}
               & Raddi                    & 1              &                        \\ \hline
Group 2        & Conic\_opacity           & 4              & 1                      \\ \hline
Group 3        & Rgbs                     & 3              & 1                      \\ \hline
Group 4        & Tiles\_touched           & 1              & n/a                    \\ \hline
\end{tabular}
}
\end{table}

\begin{table}[t!]
\caption{Texture layout for different input groups. \textmd{$n$  represents the number of Gaussians.}}
\label{tab:texture_layout}
\centering
\resizebox{\linewidth}{!}{ 
\begin{tabular}{|l|l|l|l|l|}
\hline
\textbf{Group} & \textbf{Size}       & \textbf{Width}                 & \textbf{Height}                & \textbf{Block}             \\ \hline
Group 1        & \( n \)             & \( \lceil \sqrt{\text{n}} \rceil \) & \( \lceil \text{size} / \lceil \sqrt{\text{n}} \rceil \rceil \) & \( (x, y) \) to \( (x, y) \)            \\ \hline
Group 2        & \( 2n \)            & \( \lceil \sqrt{n} \rceil \times 2 \) & \( \lceil \text{size} / \lceil \sqrt{\text{n}} \rceil \rceil \) & \( (x, y) \) to \( (x, y+1) \) \\ \hline
Group 3        & \( 12n \)           & \( \lceil \sqrt{n} \rceil \times 3 \) & \( \lceil \text{size} / \lceil \sqrt{\text{n}} \rceil \rceil \times 4 \)         & \( (x, y) \) to \( (x+3, y+4) \) \\ \hline
\end{tabular}
}
\end{table}

\noindent 
{\bf Overall Layout.} 
The overall layout is shown through Table \ref{tab:texture_layout}, where the size column indicates the total number of pixels required for each group, while the width and height columns define the dimensions of the texture dimension. The block column specifies the rectangular region (lower-left to upper-right) that stores all the information for a single Gaussian. 
The texture dimension for Group 1 is chosen to be close to the square root of the number of Gaussians to maximize the width and height while staying within texture memory limits. For Group 2, its size is twice that of Group 1, and both texture points are stored adjacent to each other to simplify index calculations. The adjacency can be either row-wise or column-wise, but in our implementation, we use row adjacency, making the width of Group 2 double that of Group 1.  

Group 3, however, is more critical, as its block layout directly impacts texture dimensions. To determine the optimal block layout, two key factors must be considered: (1) The layout should be as square as possible, and (2) it should efficiently utilize the L1 cache. Based on these criteria, we select a 3 $ \times $ 4 block dimension for Group 3.  
Overall, this configuration has two advantages:  
1)  Group 3 is three times wider than Group 1 and four times taller than Group 1. Since Group 1 is already square-like, this ensures that Group 3 maintains a near-square aspect ratio, making index calculations efficient with minimal arithmetic operations; and 2)  With a  $ 3 \times  4$ block dimension, an $ 8 \times  8$  cache block can hold two row-adjacent Gaussians along with their corresponding column-adjacent Gaussians, totaling four Gaussians, allowing for better intra-warp and local-group spatial locality. 



\subsection{Sorting Optimizations in Context of 3DGS}

\begin{figure}[t!]
    \centering
    \includegraphics[width=0.69\linewidth]{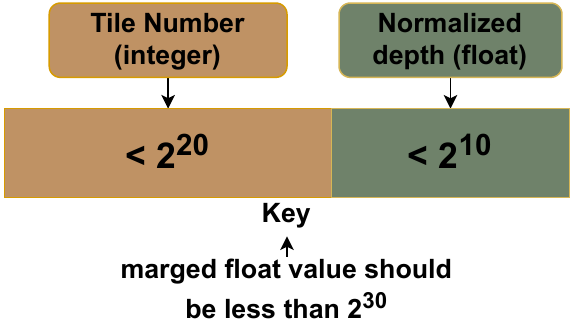}
    \caption{Key normalization illustration.}
    \label{fig:key_normalize} 
\end{figure}

The original 3D Gaussian Splatting (3DGS) method employs 64-bit integer sorting, but since our texture memory-based sorting operates on floating-point numbers, it does not support native 64-bit sorting. To address this limitation, we implement {\em key normalization}, which converts the 64-bit integer key for 3DGS to a 32-bit floating point number. Two important factors guide the decision here:  1) The key is created using the {\em tile}  number and the {\em depth},  and the keys are sorted initially by tile number and then depth;   2) Only the tile number is used in the later kernel to get the range information for each tile. So, to create a 32-bit representation, we keep 20 bits for the tile number and normalize the depth value (which is still needed during sorting) to float within the value \(2^{10}\)  (Figure \ref{fig:key_normalize}).

 Since the sorted output is subsequently used for range identification and rendering, it is crucial to store the data in a way that maximizes L1 cache efficiency. Noting that: 
 1) in the {\em identify range step}, each local work group accesses consecutive key-value pairs in the sorted dataset, and 2) during rendering, each local work group processes a 16$ \times $ 16 tile, and all threads within the group access the same range of key-value pairs. In view of this information and  
 the properties of texture cache,  we store the sorted data in a block-wise layout (Figure \ref{fig:block_sort}), ensuring that memory access patterns align with the 2D spatial locality of the L1 texture cache. Based on empirical analysis, we select a 32$\times$32 block for storing the sorted data.  

\noindent

\subsection{Efficient Tile-Based Rendering}

In the original 3D Gaussian Splatting (3DGS) approach, each local work group processes a $ 16 \times 16$  tile, with each thread responsible for computing a single pixel. After analyzing the computations involved, we observe that the operations within a tile exhibit significant data redundancy and can be optimized using Single Instruction, Multiple Data (SIMD) execution. Since all pixels within a $16 \times 16$  tile share the same set of information to compute their respective colors, we can restructure the computation so that each thread processes four pixels instead of one,  utilizing SIMD operations. 
Additionally, we also utilized loop unrolling. 
In the rendering stage, there are two loops, both of which are responsible for calculating the color for each channel. Since the number of channels is very limited (e.g., 3 channels for color images), these loops can be unrolled to eliminate loop overhead and improve  efficiency.

%% file: tex/evaluation.tex
\section{Evaluation}

\begin{figure}
    \centering
    \includegraphics[width=.40\linewidth]{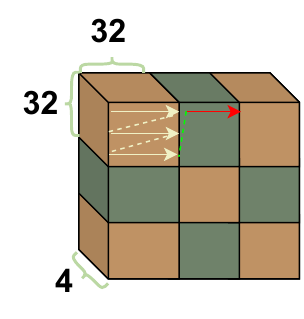}
    \caption{Block-based data storage.}
    \label{fig:block_sort} 
\end{figure} 

This section systematically assesses the performance and effectiveness of \projectname and our custom sorting pipeline. 
In all, we have the following 
objectives: 
(1) Show that our GPU-based sorting significantly outperforms existing mobile sorting techniques (or 
available implementations) in both latency and resource efficiency;
(2) Demonstrate that the full pipeline in 
\projectname outperforms 
the only other existing mobile  implementation, and  approaches  near-real-time 3D Gaussian Splatting on mobile devices without requiring additional hardware support; 
(3) Quantify how our key optimizations contribute to different components in 3DGS pipelines;
and (4) Illustrate the portability of \projectname 
across other mobile platforms.

Specifically, the evaluation includes comparisons with the state-of-the-art framework, 3dgs.cpp~\cite{3dgs.cpp}, which supports end-to-end 3DGS pipeline, featuring efficient radix sorting and rendering.
Additionally, 
since sorting is a major component of our application, 
we also compare \projectname with sorting baselines including TFLite~\cite{tensorflow2015-whitepaper} and GPUTeraSort~\cite{gputerasort}. 

\subsection{Evaluation Setup}

{\noindent\bf Testbed.} All evaluations are conducted using the off-the-shelf mobile platforms -- the Snapdragon 8 Gen 2 mobile platform, featuring an octa-core Kryo CPU and an Adreno GPU, equipped with 12 GB unified memory. For portability testing, we have used Xiaomi MI 6 (Adreno 540) and Redmi Note 10 (Mali-G57 MC2) with 8 GB and 4 GB  unified memory size, respectively. 
The baselines  used for comparison include  product-level framework TensorFlow Lite (TFLite 2.19.0)\cite{TensorFlow-Lite}, which is a widely adopted general-purpose GPU framework for mobile devices, 
and the end-to-end framework 3dgs.cpp~\cite{3dgs.cpp} we 
referred to earlier.
We have also compared our sorting with another   GPUTeraSort \cite{gputerasort}, chosen as it specifically 
targeted GPU texture-cache-aware sorting. This comparison is carried out  is through our own implementation of GPUTerasort as the implementation from the original paper is not available.  Specifically, we re-engineered its core optimizations -- such as parallel partitioning and bucket-based merging -- for compatibility with mobile hardware. In addition, we report sorting comparisons against the modern radix-sort implementation used internally by 3dgs.cpp \cite{3dgs.cpp}.
Each experiment is executed 50 times, and only the average numbers are reported, as the variance is negligible. 

\begin{figure}[t!]
    \centering
    \includegraphics[width=.90\linewidth]{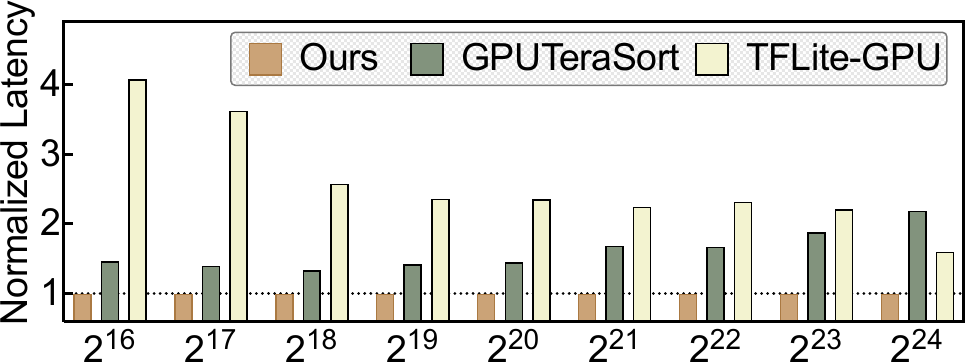}
    \caption{Latency comparison for sorting routines \textmd{with different sizes of key-value pairs (as indicated in X-axis).} \textmd{The results are normalized by Ours for readability.}}
    \label{fig:latency}
\end{figure}

{\noindent\bf Datasets.} Experiments involved multiple datasets of varying complexity: specifically, 
the 
point cloud data from the Tanks and Temples~\cite{Tanks&Temlpes}, DeepBlending\cite{DeepBlending2018} and Synthetic-Nerf\cite{mildenhall2020nerf}. Particularly, 
{\tt Train} and {\tt Truck} are  from the Tanks and Temples dataset, {\tt Playroom} is 
from DeepBlending, and {\tt Chair}, {\tt Drums}, and {\tt Lego} are from the Synthetic-NeRF dataset. 
Point cloud datasets are generated by training with the original 3DGS\cite{kerbl3Dgaussians} framework using Nvidia GPUs with 7k iterations. 
Additionally, we generate  random datasets of diverse sizes (0.5 million to 17 million key-values) to benchmark sorting efficiency.  
We omit the accuracy results since they remain the consistent among different frameworks on the same hardware. 


\subsection{Sorting Performance Comparison}
This part evaluates the performance improvements from our sorting and analyzes the underlying causes using cache profiling results.

\begin{figure}[t!]
    \centering
    \begin{subfigure}[b]{.45\linewidth}
        \centering
        \includegraphics[width=\linewidth]{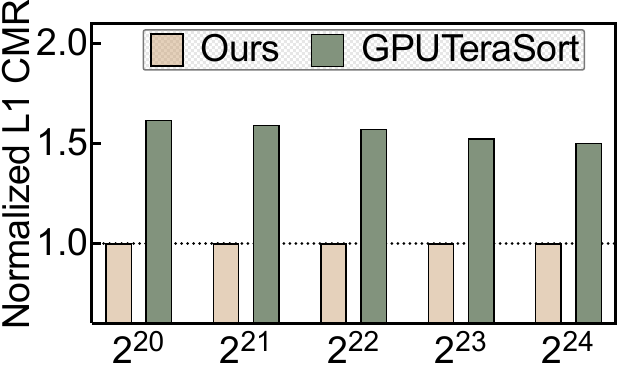}
        \caption{L1 cache miss rate}
        \label{fig:l1cache}
    \end{subfigure}
    \hfill
    \begin{subfigure}[b]{.45\linewidth}
        \centering
        \includegraphics[width=\linewidth]{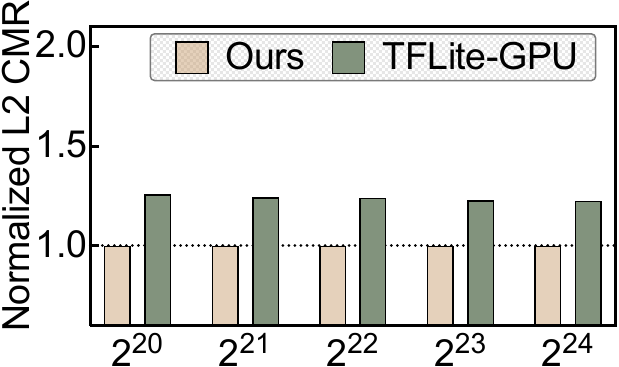}
        \caption{L2 cache miss rate}
        \label{fig:l2cache}
    \end{subfigure}
    \caption{Cache Miss Rate (CMR)  comparison for sorting. \textmd{The results are normalized by Ours for readability.}}
    \label{fig:cachemiss}
\end{figure}


\noindent 
{\bf Latency Comparison.}
Figure~\ref{fig:latency} illustrates the latency impact of our proposed sorting optimizations compared to two baseline methods (GPUTeraSort~\cite{gputerasort} and TFLite~\cite{TensorFlow-Lite}) on varied sizes of key-value pairs. For clarity, all results are normalized to our sorting approach.
Across all evaluated scenes, our sorting algorithm consistently outperforms both baselines, achieving speedups ranging from 1.5$\times$ to 4$\times$. 
These gains highlight the effectiveness of our hardware-aware optimizations, which reduce computational overhead and improve memory throughput. 

\noindent 
{\bf Cache Miss Analysis. }
Figure~\ref{fig:cachemiss} presents the cache miss rates normalized to our sorting algorithm for improved readability. To ensure a fair and meaningful comparison, we evaluate L1 cache misses against GPUTeraSort~\cite{gputerasort}, as both our method and TeraSort leverage the texture cache backed by the dedicated L1 cache on mobile GPUs. In contrast, we compare L2 cache misses with TFLite-GPU~\cite{TensorFlow-Lite}, which does not utilize the L1 texture cache and instead relies on the unified memory hierarchy.
Our sorting algorithm achieves substantial improvements in cache efficiency, reducing L1 cache misses by up to 60\% and L2 cache misses by up to 25\% over the compared baselines.  These 
improvements are primarily attributed to our hardware-aware memory access design, which promotes coalesced accesses and spatial locality.


\subsection{End-to-End Performance Comparison} 

We report results from the full pipeline, first presenting  overall latency, followed 
by memory analysis. 

\begin{figure}[t!]
    \centering
    \includegraphics[width=.90\linewidth]{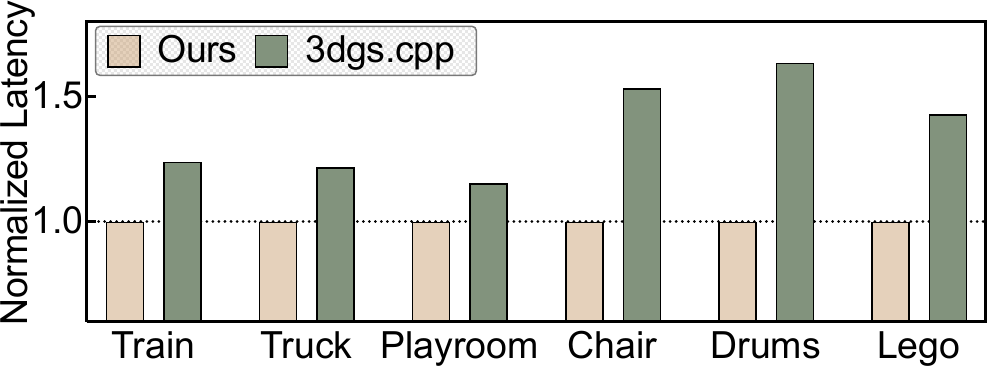}
    \caption{Overall latency (average) comparison \textmd{ for 6 datasets against 3dgs.cpp. The results are normalized by Ours.}  }
    \label{fig:endtoend}
\end{figure}

\noindent 
{\bf  Latency Comparison.}
We benchmark the end-to-end latency of our optimized 3DGS implementation against the baseline 3dgs.cpp across six scenes listed earlier.  Different scenes have different complexities and the
absolute latency of our implementations ranges from 60 ms to 600 ms, approaching near real-time 
for less complex scenes. 
The normalized latencies,  shown in Figure~\ref{fig:endtoend}, indicate 
consistent latency reductions across all cases, with 
our implementation achieved an average speedup of 
1.25$\times$.
Notably, the {\tt Drums} scene exhibits the highest speedup 
among the real-world scenes. 
This is attributed to the increased average number of visible Gaussians per tile,
which amplifies the benefits of our optimized tile-based 
renderer by improving batch execution efficiency and 
reducing per-tile overhead. 
Similarly, the {\tt Chair} and {\tt Lego} yield greater speedups than other three --  
({\tt Train},{\tt Truck}, and
{\tt  Playroom}). This turns out to be primarily 
due to their smaller model sizes,  combined with larger 
rendered image resolutions,  where our parallelism and locality 
benefits are higher. 
In contrast,  with  {\tt Train}, {\tt Playroom}, and {\tt Lego} scenes we show comparably modest improvements. This is  because they involve either fewer  Gaussians or are 
larger models, leading to  limited room for memory and execution overlap optimization.

\begin{figure}[t!]
    \centering
    \begin{subfigure}[b]{.45\linewidth}
        \centering
        \includegraphics[width=\linewidth]{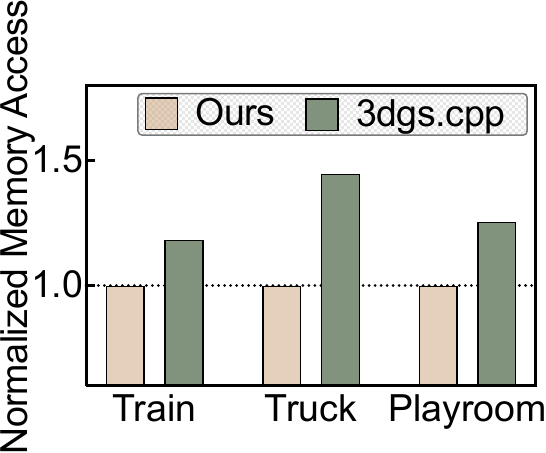}
        \caption{Memory access count.}
        \label{fig:memoryaccess}
    \end{subfigure}
    \hfill
    \begin{subfigure}[b]{.45\linewidth}
        \centering
        \includegraphics[width=\linewidth]{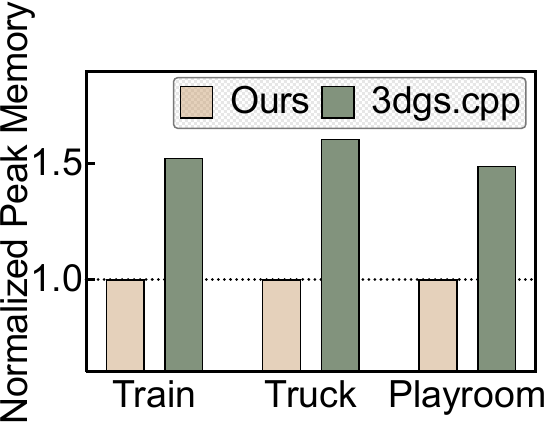}
        \caption{Peak memory consumption.}
        \label{fig:peakmemory}
    \end{subfigure}
    \caption{Overall memory usage comparison. The results are normalized by ours for readability.}
    \label{fig:memory}
\end{figure}

\noindent 
{\bf  Memory Efficiency Comparison.}
We evaluate memory efficiency through detailed runtime profiling, focusing on both memory access volume and peak memory usage. 
The results for {\tt Chair}, {\tt Drums}, and {\tt Lego} show similar trends  and 
are omitted. 
As shown in Figure~\ref{fig:memoryaccess} and Figure~\ref{fig:peakmemory}, our optimized implementation reduces total memory accesses by 25\% and peak memory usage by 20\% on average, compared to the baseline. These improvements stem from our  our optimizations involving variable packing, block-wise data layout, 
-- they all  reduce both the data movement and memory footprint.
Among the evaluated scenes, {\tt Truck} exhibits the highest reduction in memory accesses. This is because the number of tile-Gaussian pairs in {\tt Truck} is relatively small compared to the total number of Gaussians, enabling more efficient indexing and pruning during rendering. In contrast, {\tt Playroom} demonstrates a smaller memory reduction. This is due to  inefficiencies in the 
underlying  base sorting scheme, which pads the number of key-value pairs to the nearest power-of-two. 
For {\tt Playroom}, the actual number of pairs is substantially lower than the next power-of-two threshold, resulting in underutilized texture memory. 

\begin{table*}[t!]
\caption{Latency comparison (ms) across different 3DGS pipelines on different datasets and scenes. \textmd{``w/o'' refers to without that optimization. ``sz'' refers to image size and ``n'' refers to the number of 3d gaussians. ``dl'' shorts for optimal data layout. ``vp'' and ``ec'' stand for variable packing and execution fusion, respectively. All versions in this table employ our optimized texture-memory-aware sorting approach, hence  sorting latency is almost unchanged among all versions.}}
\label{tab:performance_comparison}
\centering
\setlength{\tabcolsep}{3pt} 
\resizebox{\textwidth}{!}{
\begin{tabular}{|l|l|c|c|c|c|c|c|c|c|}
\hline
\textbf{Scene} & \textbf{Approach} & \textbf{Speedup} & \textbf{End to End} & \textbf{Preprocess} & \textbf{Scan} & \textbf{Duplicate with tiles} & \textbf{Sorting} & \textbf{Identify Range} & \textbf{Render} \\
\hline

\multirow{4}{*}{\makecell{Train\\sz : 980$\times$545\\n : 741,295}} 
& \textbf{Ours} & \multirow{4}{*}{\textbf{1.21$\times$}} & \textbf{299} & \textbf{4.60} & \textbf{2.70} & \textbf{3.19} & \textbf{267} & \textbf{0.47} & \textbf{20.9} \\
& Ours w/o dl &  & 310 & 5.40 & 2.67 & 3.19 & 267 & 0.69 & 30.6 \\
& Ours w/o vp + dl &  & 326 & 8.60 & 2.67 & 5.60 & 267 & 0.69 & 40.8 \\
& Ours w/o vp + dl + ec &  & 361 & 8.60 & 2.67 & 5.60 & 267 & 0.69 & 76.0 \\
\hline

\multirow{4}{*}{\makecell{Truck\\sz : 979$\times$546\\n : 1,689,804}} 
& \textbf{Ours} & \multirow{4}{*}{\textbf{1.27$\times$}} & \textbf{314} & \textbf{6.80} & \textbf{6.94} & \textbf{3.04} & \textbf{270} & \textbf{0.46} & \textbf{25.8} \\
& Ours w/o dl &  & 342 & 10.7 & 6.94 & 3.04 & 271 & 0.64 & 49.4 \\
& Ours w/o vp + dl &  & 370 & 25.4 & 6.94 & 5.49 & 271 & 0.64 & 60.6 \\
& Ours w/o vp + dl + ec &  & 400 & 25.4 & 6.94 & 5.49 & 271 & 0.64 & 90.5 \\
\hline

\multirow{4}{*}{\makecell{Playroom\\sz : 1264$\times$832\\n : 1,491,851}} 
& \textbf{Ours} & \multirow{4}{*}{\textbf{1.18$\times$}} & \textbf{584} & \textbf{5.60} & \textbf{6.10} & \textbf{5.30} & \textbf{506} & \textbf{0.83} & \textbf{59.5} \\
& Ours w/o dl &  & 619 & 9.20 & 6.10 & 5.30 & 506 & 1.83 & 90.3 \\
& Ours w/o vp + dl &  & 648 & 20.2 & 6.10 & 7.90 & 506 & 1.83 & 105 \\
& Ours w/o vp + dl + ec &  & 690 & 20.2 & 6.10 & 7.90 & 506 & 1.83 & 148 \\
\hline
\end{tabular}
}

\end{table*}

\begin{figure}[t!]
    \centering
    \includegraphics[width=.90\linewidth]{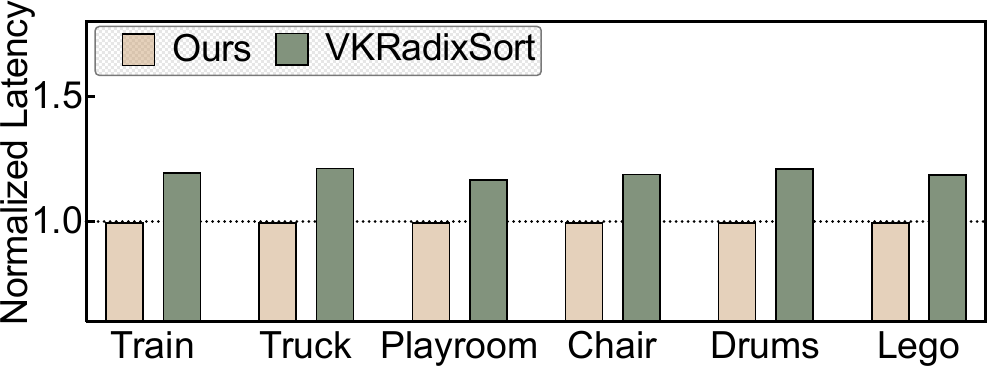}
    \caption{Comparison of  (average) end-to-end  latency of our implementation against VKRadixSort for 6 datasets.  The results are normalized by Ours for readability.}
    \label{fig:sortwithvk}
    \vspace{-0.5em}
\end{figure}

\subsection{Benefits of Other Optimizations} 

Table~\ref{tab:performance_comparison} compares the performance of our full pipeline (``Ours'') with its ablated variants across three representative scenes: {\tt Train} and {\tt Truck} from the Tanks and Temples dataset, and {\tt Playroom} from Deep Blending.
We evaluate the impact of disabling three key optimizations: optimal data layout (dl), variable packing (vp), and execution fusion (ec). All versions include our optimized sorting implementation, 
as we have already studied benefits from sorting. 
Our full version achieves the best end-to-end latency, 
with up to a 1.27$\times$ improvement over the least 
optimized variant (but still with optimized sorting). 
The performance gains are most prominent in the rendering  and preprocessing stages because these  kernels heavily depend on a huge number of data read/write operations.
Despite {\tt Truck} and {\tt Playroom} having similar numbers of 3D Gaussians,
the  sorting cost in {\tt Playroom} is nearly double that of Truck.
This is attributed to the significantly larger number of Gaussian-tile pairs in {\tt Playroom}, which directly increases the sorting workload.
We further observe that data layout optimizations influence multiple stages: preprocess, identify range, and render. 
This is because preprocessing handles spherical harmonics (Shs) and writes to a blockwise layout, which is then consumed downstream. 
The layout also improves spatial locality during rendering and range identification. Data layout selection  (dl) contributes 1.05$\times$ to 1.07$\times$ speedup on the selected datasets. 
Variable packing, applied to both inputs and outputs of the preprocessing stage, also impacts Duplicate with Tiles and render, since packed buffers are reused across these stages. 
When disabled, the memory footprint and processing time increase due to less efficient data movement and reuse. 
This optimization  adds 1.03$\times$ to 1.05$\times$ speedup. 
Lastly, the execution fusion (ec) optimization is specific to the render stage. Disabling it only affects rendering time, since this optimization is limited to kernel-level fusion and instruction scheduling within that stage. It offers 1.06$\times$ to 1.1$\times$ speedup.

Apart from the above optimizations, we also compare our sorting implementation with VKRadixSort\cite{vkradixsort}, which is a widely used radix-sort implementation available for mobile GPUs. 
As shown in Figure \ref{fig:sortwithvk}, our approach consistently achieves lower  latency than VKRadixSort.
Specifically, Figure \ref{fig:sortwithvk} shows that the sorting algorithm alone achieves a performance improvement of 1.10$\times$ to 1.15$\times$ over VKRadixSort.


\begin{figure}[t!]
    \centering
    \begin{subfigure}[b]{.45\linewidth}
        \centering
        \includegraphics[width=\linewidth]{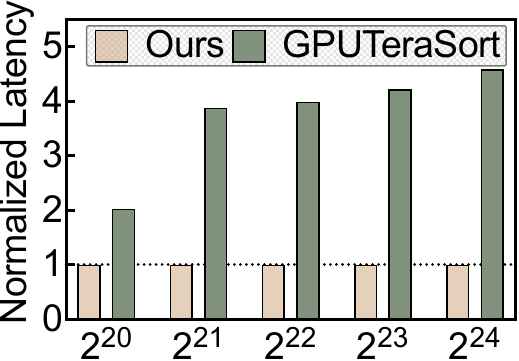}
        \caption{XiaoMi 6}
        \label{fig:mi6}
    \end{subfigure}
    \hfill
    \begin{subfigure}[b]{.45\linewidth}
        \centering
        \includegraphics[width=\linewidth]{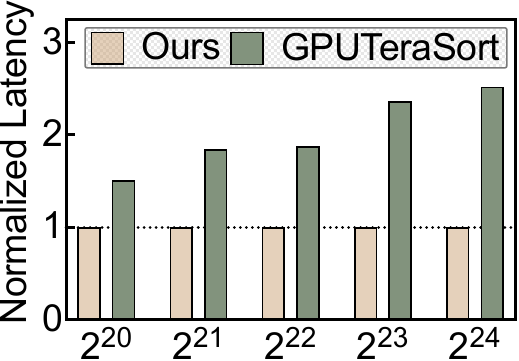}
        \caption{Redmi Note 10}
        \label{fig:redminote10}
    \end{subfigure}
    \caption{Sorting time comparison on 2 older devices.}
    \label{fig:sortold}
    \vskip -0.1 cm
\end{figure}

\subsection{Portability}\label{sec:eva-portability}
Figure~\ref{fig:sortold} reports the execution latency of our sorting implementation and 
compares to the GPUTeraSort\cite{gputerasort} on two additional mobile devices--Xiaomi MI 6 (Adreno 540) and Redmi Note 10 (Mali-G57 MC2). 
This experiment focuses on demonstrating portability by selecting the sorting implementation, as the baseline for the full 3DGS could not be run on older platforms.
Our implementation consistently delivers stable performance across both devices, particularly on the older-generation MI 6. 
Our optimizations greatly reduce intermediate memory usage and computational complexity, ensuring robustness -- especially important for devices with limited memory bandwidth and processing power.

%% file: tex/related.tex
\section{Related Work} 





\compactparagraph{End-to-end 3DGS Acceleration.}
To address efficiency issues for 3DGS, a number of end-to-end acceleration strategies have been proposed. 
{Mini-Splatting}~\cite{Fang2024MiniSplat} represents one early effort to constrain model size by reorganizing Gaussians in the scene. 
{LightGaussian}~\cite{LightGaussian2023} takes a complementary approach by aggressively compressing the 3DGS representation: it prunes away globally least significant Gaussians and quantizes remaining parameters.
{EfficientGS}~\cite{Chen2023EfficientGS} focuses on reducing per-Gaussian complexity, for instance by retaining only low-order spherical harmonic coefficients for most Gaussians and increasing the coefficient order or density only when needed.
{EagleS}~\cite{girish2025eagles} introduces a novel Gaussian pruning strategy that leverages a learned saliency map to identify and remove redundant Gaussians.
{Morgenstern et al.} propose a compressed 3D scene that ~\cite{morgenstern2024compact} primarily addresses compact data representation through structured 2D grids rather than runtime sorting efficiency. 
These methods collectively demonstrate that end-to-end performance improvements are possible by simplifying or restructuring the entire 3DGS pipeline. 
However, they are primarily designed for desktop GPUs and rely on high memory bandwidth, shared memory, and large-scale parallelism. 
Such assumptions do not hold on mobile GPUs, where memory and compute resources are significantly constrained. 
  The sort-free methods, OIT rendering~\cite{hou2024sort} and StochasticSplats~\cite{kheradmand2025stochasticsplats}, 
  effectively remove sorting overhead but introduce significant computational complexities. 
  StochasticSplats relies on expensive multi-sample Monte Carlo estimation, 
  and OIT rendering adds considerable per-fragment computation overhead. 
  In our experiments, these methods typically require substantial computational resources and are not able to directly apply to resource-constrained mobile devices used in our experiments.

\compactparagraph{Memory Optimizations for 3DGS.}
Another line of work focuses on algorithm-level optimizations to reduce 3DGS memory usage and runtime cost. 
{Taming~3DGS}~\cite{mallick2024taming} adopts a saliency-based approach by selectively densifying or pruning Gaussians using image-space gradient metrics. 
Most recently, {GaussianSpa}~\cite{zhang2024gaussianspa} formulates a sparsity-constrained optimization problem, applying an alternating minimization framework to jointly optimize reconstruction and sparsification. 
{3DGS-LM}~\cite{Hoel2024} replaces the Adam optimizer with a tailored Levenberg–Marquardt solver, achieving faster convergence during reconstruction through second-order updates and view-consistent gradient fusion.
Orthogonal to pruning, other methods leverage quantization and compression to reduce per-Gaussian storage overhead. 
{CompGS}~\cite{navaneet2023compact3d} applies vector quantization with codebooks for spatial and color attributes. 
{RDO-Gaussian}~\cite{Papantonakis2024} combines pruning and entropy-constrained quantization in a rate-distortion framework, achieving efficient storage with bounded reconstruction loss. 
{EfficientGS}~\cite{Chen2023EfficientGS} reduces spherical harmonic (SH) order adaptively, compressing color representation without compromising rendering quality.
While these works have made meaningful progress in simplifying Gaussian representations and accelerating training, our optimization is orthogonal to these approaches.  Combining  our sorting and memory optimizations with these compression techniques can be promosing directions for future work. 

\compactparagraph{GPU Sorting Optimization.}
While extensively studied on desktop GPUs, sorting remains a bottleneck on mobile GPUs due to its irregular memory access pattern.
Classical GPU sorting methods —- such as parallel radix sort~\cite{Satish2009}, bitonic sort~\cite{Batcher1968}, and sample sort -- leverage warp-level parallelism and shared memory, achieving high throughput in large-scale settings. 
These methods exemplify how cache-aware GPU designs can scale sorting throughput with large input sizes.
However, these designs assume access to large, shared memory regions and high-bandwidth memory interfaces. 
Mobile GPUs, by contrast, employ tile-based deferred rendering with highly localized memory and lower peak throughput. 
As a result, GPU sorting algorithms optimized for desktop architectures often underperform or become infeasible on mobile platforms. 


%% file: tex/conclusion.tex
\section{Conclusions and Future Work} 

In this paper, we presented a novel optimization framework for accelerating 3DGS applications on mobile GPUs. Our approach specifically targets efficient computation utilizing the 2D texture cache on mobile GPU architectures. 
The primary innovation is a novel sorting algorithm that significantly reduces cache misses by optimizing data movement and layout, along with enhanced data placement strategies for greater overall efficiency. 
Extensive evaluations show that our end-to-end implementation achieves up to 1.6$\times$ performance improvement compared to baseline implementations, with particularly significant gains in sorting kernel operations by up to 4.1$\times$.
Compared to the alternative 3dgs frameworks and sorting solutions, our solution provides novel insights into exploiting 2D memory characteristics.  
The work presented here can be extended in multiple directions. One area can be revisiting the 
proposed optimizations to support adaptive resolution models. Another area can be 
exploring sorting and full application enhancements for other mobile architectures, e.g., mobile NPUs or DSP 
chips. 



%% file: tex/ack.tex
\section*{Acknowledgment}

The authors want to extend their appreciation to the anonymous reviewers 
for their valuable and thorough feedback, which helped improve the paper. 
This work was supported in part by the National Science Foundation (NSF) under the awards of
CCF-2428108, 
OAC-2403090, CNS-2341378, and CCF-2333895. 
Any errors and opinions are not those of the NSF and are attributable solely to the author(s).